\documentclass[11pt,letterpaper]{article}
\usepackage{emnlp2017}
\usepackage{times}
\usepackage{latexsym}

\usepackage{graphicx}
\usepackage{amsmath}
\usepackage{multirow}
\usepackage{subfigure}
\usepackage{amssymb}
\usepackage{url}

\emnlpfinalcopy

\def\emnlppaperid{60}

\title{Content-Based Table Retrieval for Web Queries}

\author{Zhao Yan$^\dag$\Thanks{\ \ Contribution during internship at \mbox{Microsoft Research}.}\ , Duyu Tang$^\ddag$ , Nan Duan$^\ddag$ , Junwei Bao$^{+*}$ ,\\ \textbf{Yuanhua Lv$^\S$ , Ming Zhou$^\ddag$ ,	Zhoujun Li$^\dag$} \\
	$^\dag$Beihang University 
	\hspace{0.5cm} $^\ddag$Microsoft Research, Beijing, China \\
	$^+$Harbin Institute of Technology 
	$^\S$Microsoft AI and Research, Sunnyvale CA, USA\\
	$^\dag${\tt \{yanzhao, lizj\}@buaa.edu.cn} \hspace{1.3cm}
	$^+${\tt baojunwei001@gmail.com} \\
	$^\ddag$$^\S${\tt \{dutang, nanduan, yuanhual, mingzhou\}@microsoft.com}
	\\
} 
\date{}

\begin{document}

\maketitle

\begin{abstract}
  Understanding the connections between unstructured text and semi-structured table is an important yet neglected problem in natural language processing.
  In this work, we focus on content-based table retrieval. Given a query, the task is to find the most relevant table from a collection of tables.
  Further progress towards improving this area requires
  powerful models of semantic matching and richer training and evaluation resources.
  To remedy this, we present a ranking based approach, and implement both carefully designed features and neural network architectures to measure the relevance between a query and the content of a table.
  Furthermore, we release an open-domain dataset that includes 21,113 web queries for 273,816 tables.
  We conduct comprehensive experiments on both real world and synthetic datasets.
  Results verify the effectiveness of our approach and present the challenges for this task.
\end{abstract}


\section{Introduction}
Table\footnote{\url{https://en.wikipedia.org/wiki/Table\_(information)} } is a special and valuable information that could be found almost everywhere from the Internet.
We target at the task of content-based table retrieval in this work.
Given a query, the task is to find the most relevant table from a collection of tables.
Table retrieval is of great importance for both natural language processing and information retrieval.
On one hand, it could improve existing information retrieval systems.
The well-organized information from table, such as product comparison from different aspects and flights between two specific cities, could be used to directly respond to web queries.
On the other hand, the retrieved table could be used as the input for question answering \cite{pasupat2015wtq}.

Unlike existing studies in database community \cite{VLDB2008GG,CIDR2015GG} that utilize surrounding text of a table or pagerank score of a web page, we focus on making a thorough exploration of table content in this work. 
We believe that content-based table retrieval has the following challenges.
The first challenge is how to effectively represent a table, which is semi-structured and includes many aspects such as headers, cells and caption.
The second challenge is how to build a robust model that measures the relevance between an unstructured natural language query and a semi-structured table.
Table retrieval could be viewed as a multi-modal task because the query and the table are of different forms.
Moreover, to the best of our knowledge, there is no publicly available dataset for table retrieval.
Further progress towards improving this area requires richer training and evaluation resources.

To address the aforementioned challenges, we develop a ranking based approach.
We separate the approach into two cascaded steps to trade-off between accuracy and efficiency.
In the first step, it finds a small set (e.g. 50 or 100) of candidate tables using a basic similarity measurement.
In the second step, more sophisticated features  are used to measure the relevance between the query and each candidate table.
We implement two types of features, including manually designed features inspired by expert knowledge and neural network models jointly learned from data.
Both strategies take into account the relevance between query and table at different levels of granularity.
We also introduce a new dataset \textit{WebQueryTable} for table retrieval. It includes 21,113 web queries from search log, and 273,816 web tables from Wikipedia.

We conduct comprehensive experiments on two datasets, a real world dataset introduced by us, and a synthetic dataset WikiTableQuestions \cite{pasupat2015wtq} which has been widely used for table-based question answering.
Results in various conditions show that neural network models perform comparably with carefully designed features, and combining them both could obtain further improvement.
We study the influence of each aspect of table for table retrieval, and show what depth of table understanding is required to do well on this task.
Results show the difference between question and web query, and present future challenges for this task.

This paper has the following contributions.
We develop both feature-based and neural network based approaches, and conduct thorough experiments on real world and synthetic datasets.
We release an open-domain dataset for table retrieval.

\section{Task Definition}

We formulate the task of table retrieval in this section.
Given a query $q$ and a collection of tables $T=\{t_1, ..., t_N\}$, the goal of table search is to find a  table $t_i$ that is most relevant to $q$.

\begin{figure}[h]
	\centering
	\includegraphics[width=0.49\textwidth]
	{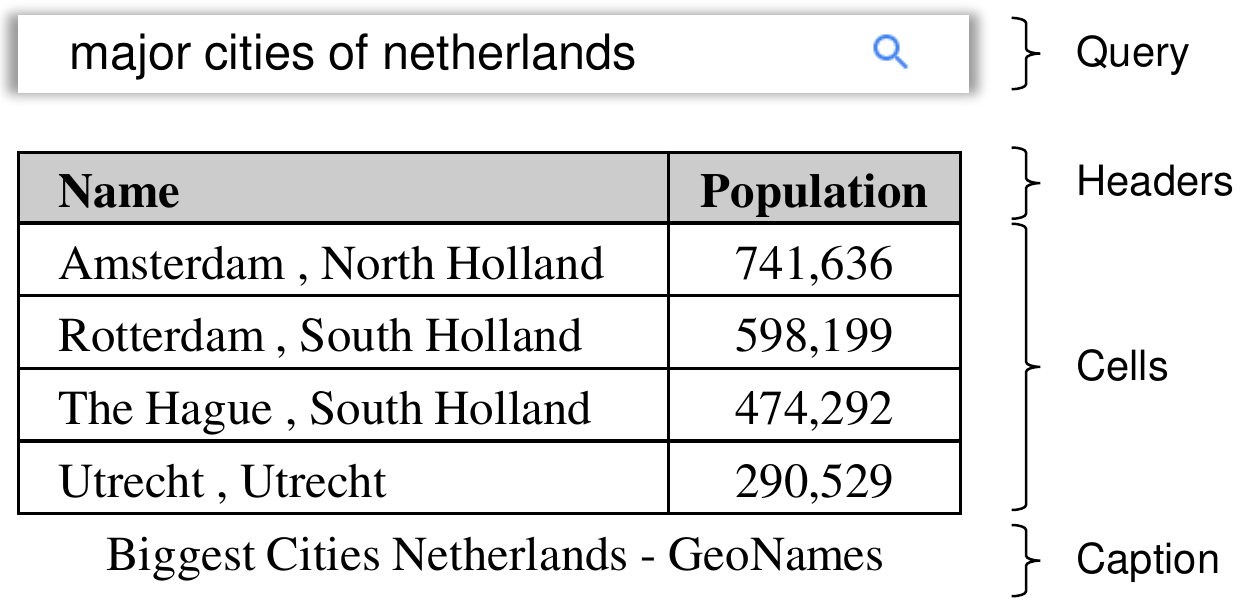}
	\caption{A example of query-table pair.}
\end{figure}

Typically, a query $q$ is a natural language expression that consists of a list of words, such as ``\textit{major cities of netherlands}''.
A table $t$ is a set of data elements arranged by vertical columns and horizontal rows.
Formally, we define a table as a triple $t=\{headers,\ cells,\ caption\}$ that consists of three aspects.
A table could have multiple $headers$, each of which indicates the property of a column and could be used to identify a column.
A table could have multiple $cells$,  each of which is a unit where a row and a column intersects.
A table could have a $caption$, which is typically an explanatory text about the table.
Figure 1 gives an example to illustrate different aspects of a table.

It is helpful to note that tables from the web are not always ``regular''.
We regard a table as a ``regular'' table if it contains header, cell and caption, and the number of cells in each row is equal to the number of header cells.
In this work, we make a comprehensive study of table retrieval on regular tables, and would like to release benchmark datasets of good quality.
It is trivial to implement heuristic rules so as to convert the irregular tables to regular one, so we leave it to the future work.

\section{Approach Overview}
In this section, we give an overview of the proposed approach.
To build a system with high efficiency, we separate the task into two cascaded modules, including candidate table retrieval and table ranking.
Candidate table retrieval aims to find a small set of tables, such as 50 or 100.
These candidate tables will be further used in the table ranking step, which uses more sophisticated features to measure the relevance between a query and a table.
In the following subsections, we will give the work-flow of candidate table retrieval and table ranking.
The detailed feature representation will be described in the next section.

\subsection{Candidate Table Retrieval}

Candidate table retrieval aims to get a small candidate table set from the whole table set of large scale, which is hundreds of thousands in our experiment.
In order to guarantee the efficiency of the searching process, we calculate the similarity between table and query with Okapi BM25 \cite{bm25}, which is computationally efficient and has been successfully used in information retrieval.
Specifically, we represent a query as bag-of-words, and represent table with plain text composed by the words from caption and headers. 
Given a query $q = {x_1, x_2, ..., x_n}$, a table $t$ and the whole table set $T$, the BM25 score of query $q$ and table $t$ is calculated as follows.

\begin{small}
	\begin{equation}
	BM25(q, t) \\
	= \sum_{i=1}^{n} idf(x_{i}) \frac{tf(x_{i}, t) \cdot (k_1+1)}{tf(x_{i}, T) + k_1 (1-b+b \frac{|t|}{avg_{tl}})}   \nonumber
	\end{equation}
\end{small}
where $tf(x_{i}, t)$ is the term frequency of word $x_i$ in $t$, $idf(x_i)$ is its inverse document frequency, $avg_{tl}$ is the average sequence length in the whole table set $T$, and $k_1$ and $b$ are hyper-parameters.

\subsection{Table Ranking}
The goal of table ranking is to rank a short list of candidate tables by measuring the relevance between a query and a table. 
We develop a feature-based approach and a neural network approach, both of them effectively take into account the structure of table.
The details about the features will be described in next section.
We use each feature to calculate a relevance score, representing the similarity between a query and a table from some perspective.
Afterwards, we use LambdaMART \cite{mart2010l2r}, a successful algorithm for solving real world ranking problem, to get the final ranking score of each table.\footnote{We also implemented a ranker with linear regression, however, its performance was obviously worse than LambdaMART in our experiment.}
The basic idea of LambdaMART is that it constructs a forest of decision trees, and its output is a linear combination of the results of decision trees.
Each binary branch in a decision tree specifies a threshold to apply to a single feature, and each leaf node is real value.
Specifically, for a forest of $N$ trees, the relevance score of a query-table pair is calculated as follow,
\begin{equation}
s(q,t)
= \sum_{i=1}^{N} w_i tr_i(q,t)   \nonumber
\end{equation}
where $w_i$ is the weight associated with the $i$-th regression tree, and $tr_i( \cdot )$ is the value of a leaf node obtained by evaluating $i$-th tree with features $\left[ f_1(q,t), ... ,f_K(q,t) \right]$.
The values of $w_i$ and the parameters in $tr_i(\cdot)$ are learned with gradient descent during training.

\section{Matching between Query and Table}
Measuring the relevance between a query and a table is of great importance for table retrieval.
In this section, we present carefully designed features and neural network architectures for matching between a query and a table.

\subsection{Matching with Designed Features}
\label{section:features}
We carefully design a set of features to match query and table from word-level, phrase-level and sentence-level, respectively.
The input of a feature function are two strings, one query string $q$ and one aspect string $t_a$.
We separately apply each of the following features to each aspect of a table, resulting in a list of feature scores.
As described in Section 2, a table has three aspects, including headers, cells and caption.
We represent each aspect as word sequence in this part.

(1) \textbf{Word Level.} We design two word matching features $f_{wmt}$ and $f_{mwq}$. The intuition is that a query is similar to an aspect of table if they have a large amount of word overlap.
$f_{wmt}$ and $f_{wmq}$ are calculated based on number of words shared by $q$ and $t_a$.
They are also normalized with the length of $q$ and $t_a$, calculated as follows,

\begin{align}
	f_{wmt}(t_{a}, q)&=\frac{\sum_{w \in t_{a}} \delta(w, q) \cdot idf(w)}{\sum_{w' \in t_{a}} idf(w')} \nonumber \\
	f_{wmq}(t_{a}, q)&=\frac{\sum_{w \in t_{a}} \delta(w, q) \cdot idf(w)}{\sum_{w' \in q} idf(w')} \nonumber
\end{align}

where $idf(w)$ denotes the inverse document frequency of word $w$ in $t_{a}$.
$\delta(y_j, q)$ is an indicator function which is equal to 1 if $y_j$ occurs in $q$, and 0 otherwise.
Larger values of $f_{wmt}(\cdot)$ and $f_{wmq}(\cdot)$ correspond to larger amount of word overlap between $t_a$ and $q$.

(2) \textbf{Phrase Level.}
We design a paraphrase-based feature $f_{pp}$ to deal with the case that a query and a table use different expressions to describe the same meaning.
In order to learn a strong and domain-independent paraphrase model, we leverage existing statistical machine translation (SMT) phrase tables.
A phrase table is defined as a quadruple, namely $PT = \{ \langle src_i,trg_i, p(trg_i|src_i), p(src_i|trg_i) \rangle\}$,
where $src_i$ (or $trg_i$) denotes a phrase, in source (or target) language,
$p(trg_i|src_i)$ (or $p(src_i|trg_i)$) denotes the translation probability from $srg_i$ (or $trg_i$) to $trg_i$ (or $src_i$).
We use an existing SMT approach \cite{koehn2003PP} to extract a phrase table $PT$ from a bilingual corpus.
Afterwards, we use $PT$ to calculate the relevance between a query and a table in paraphrase level.
The intuition is that, two source phrases that are aligned to the same target phrase tend to be paraphrased.
The phrase level score is calculated as follows, where $N$ is the maximum n-gram order, which is set as 3, and $src_{i,n}^{a_t}$ and $src_{j,n}^{q}$ are the phrase in $t_{a}$ and $q$ starts from the $i$-th and $j$-th word with the length of $n$,
and $i \in \left\lbrace1,...,|t_a|-n+1\right\rbrace$ and $j \in \left\lbrace 1,...,|q|-n+1\right\rbrace $.

\begin{small}
	\begin{eqnarray}
	f_{pp}(t_{a},q)= \frac{1}{N}\sum_{n=1}^N \frac{\sum_{i,j} score(src_{i,n}^{t_q}, src_{j,n}^{q})}{|t_a|-N+1} \nonumber \\
	score(src_x;src_y)=\sum_{PT}p(tgt_k|src_x) \cdot p(src_y|tgt_k) \nonumber
	\end{eqnarray}
\end{small}

(3) \textbf{Sentence Level.}
We design features to match a query with a table at the sentence level.
We use CDSSM \cite{shen2014CDSSM}, which has been successfully applied in text retrieval.
The basic computational component of CDSSM is sub-word, which makes it very suitable for dealing the misspelling queries in web search.
The model composes sentence vector from sub-word embedding via convolutional neural network.
We use the same model architecture to get query vector and table aspect vector, and calculate their relevance with cosine function.
\begin{eqnarray}
f_{s1}(t_a, q)=cosine(cdssm(t_a), cdssm(q)) \nonumber
\end{eqnarray}
We train model parameters on WikiAnswers dataset \cite{Fader13WikiAnswer}, which contains almost 12M question-similar question pairs.
In addition, since vector average is an intuitive way to compute sentence vector and does not induce additional parameters, we calculate another relevance score by
representing a query and a table aspect with element-wise vector average.
We use a publicly available word embedding which is released by \newcite{mikolov2013w2v}.
\begin{eqnarray}
f_{s2}(t_a, q)=cosine(vec\_avg(t_a), vec\_avg(q)) \nonumber
\end{eqnarray}

\subsection{Matching with Neural Networks}
\label{section:nn}
We present neural network models for matching a query with a table.
As a table includes different aspects such as headers, cells and caption, we develop different strategies to measure the relevance between a query and a table from different perspectives.
In this subsection, we first describe the model to compute query representation, and then present the method that measures the relevance between a query and each aspect.

A desirable query representation should be
sensitive to word order as reversing or shuffling the words in a query might result in totally different intention.
For example, ``\textit{list of flights london to berlin}" and ``\textit{list of flights berlin to london}" have different intentions.
We use recurrent neural network (RNN) to map a query of variable length to a fixed-length vector.
To avoid the problem of gradient vanishing, we use gated recurrent unit (GRU) \cite{cho-EtAl:2014:EMNLP2014} as the basic computation unit, which adaptively forgets the history and remembers the input, and has proven to be effective in sequence modeling \cite{chung2014empirical}.
It recursively transforming current word vector $e^q_t$ with the output vector of the previous step $h_{t-1}$.

\begin{eqnarray}
&z_i &= \sigma(W_{z}e^q_{i} + U_{z}{h}_{i-1}) \nonumber \\
&r_i &= \sigma(W_{r}e^q_{i} + U_{r}{h}_{i-1}) \nonumber \\
&\widetilde{h}_i &= \tanh(W_{h}e^q_{i} + U_{h}(r_i \odot {h}_{i-1})) \nonumber \\
&{h}_{i} &= z_i \odot \widetilde{h}_i + (1-z_i) \odot {h}_{i-1}  \nonumber
\end{eqnarray}
where $z_i$ and $r_i$ are update and reset gates of GRU.
We use a bi-directional RNN to get the meaning of a query from both directions, and use the concatenation of two last hidden states as the final query representation $v_q=[ \overrightarrow{h}_n , \overleftarrow{h}_n ]$.

A table has different types of information, including headers, cells and caption. We develop different mechanisms to match the relevance between a query and each aspect of a table.
An important property of a table is that randomly exchanging two rows or tow columns will not change the meaning of a table \cite{vinyals2015order}.
Therefore, a matching model should ensure that exchanging rows or columns will result in the same output.
We first describe the method to deal with headers.
To satisfy these conditions, we represent each header as an embedding vector, and regard a set of header embeddings as external memory $M_h \in \mathbb{R}^{k \times d}$, where $d$ is the dimension of word embedding, and $k$ is the number of header cells.
Given a query vector $v_q$, the model first assigns a probability $\alpha_i$ to each memory cell $m_i$, which is a header embedding in this case.
Afterwards, a query-specific header vector is obtained through weighted average \cite{Bahdanau2015,Sukhbaatar2015e2emn}, namely $v_{header} = \sum_{i=1}^{k}\alpha_i m_i$, where $\alpha_i \in [0,1]$ is the weight of $m_i$ calculated as below and $\sum_{i} \alpha_i = 1$.
\begin{equation}
\alpha_i = \frac{exp(tanh(W [m_i; v_q] + b))}{\sum_{j=1}^k exp(tanh(W [m_j; v_q] + b))}\nonumber
\end{equation}
Similar techniques have been successfully applied in table-based question answering \cite{pengcheng2015,neelakantan2015neural}.
Afterwards, we feed the concatenation of $v_q$ and $v_{header}$ to a linear layer followed by a $softmax$ function whose output length is 2. We regard the output of the first category as the relevance between query and header. We use $NN_1()$ to denote this model.
\begin{eqnarray}
f_{nn}(header, q)=NN_{1}(M_{h}, v_{q}) \nonumber
\end{eqnarray}

Since headers and cells have similar characteristics, we use a similar way to measure the relevance between a query and table cells.
Specifically, we derive three memories $M_{cel}$, $M_{row}$ and $M_{col}$ from table cells in order to match from cell level, row level and column level.
Each memory cell in $M_{cel}$ represents the embedding of a table cell.
Each cell in $M_{row}$ represent the vector a row, which is computed with weighted average over the embeddings of cells in the same row. We derive the column memory $M_{col}$ in an analogous way.
We use the same module $NN_1()$ to calculate the relevance scores for these three memories.
\begin{eqnarray}
f_{nn}(cell, q)&=&NN_{1}(M_{cel}, v_{q}) \nonumber \\
f_{nn}(column, q)&=&NN_{1}(M_{col}, v_{q}) \nonumber \\
f_{nn}(row, q)&=&NN_{1}(M_{row}, v_{q}) \nonumber
\end{eqnarray}

Since a table caption is typically a descriptive word sequence. We model it with bi-directional GRU-RNN, the same strategy we have used for modeling the query.
We concatenate the caption vector $v_{cap}$ with $v_{q}$, and feed the results to a linear layer followed by $softmax$.
\begin{eqnarray}
f_{nn}(caption, q)=NN_{2}(v_{cap}, v_{q}) \nonumber
\end{eqnarray}

We separately train the parameters for each aspect with back-propagation.
We use negative log-likelihood as the loss function.\footnote{We also implemented a ranking based loss function $ max(0, 1 - f_{nn}(t_a,q) + f_{nn}(t_a^*,q))$, but it performed worse than the negative log-likelihood in our experiment.}
\begin{equation}
loss = -\frac{1}{|D|}\sum_{(t_a, q) \in D} \log(f_{nn}(t_a,q)) \nonumber
\end{equation}

\section{Experiment}
We describe the experimental setting and analyze the results in this section.
\subsection{Dataset and Setting}

To the best of our knowledge, there is no publicly available dataset for table retrieval.
We introduce \textbf{WebQueryTable}, an open-domain dataset consisting of query-table pairs.
We use search logs from a commercial search engine to get a list of queries that could be potentially answered by web tables.
Each query in query logs is paired with a list of web pages, ordered by the number of user clicks for the query.
We select the tables occurred in the top ranked web page, and ask annotators to label whether a table is relevant to a query or not.
In this way, we get 21,113 query-table pairs.
In the real scenario of table retrieval, a system is required to find a table from a huge collection of tables.
Therefore, in order to enlarge the search space of our dataset, we extract 252,703 web tables from Wikipedia and regard them as searchable tables as well. 
Data statistics are given in Table \ref{table:statistic}.

\begin{table}[h]
	\begin{small}
		\centering
		\begin{tabular}{p{2.67cm}|c|c}
			\hline
			& \textbf{WQT dataset} & \textbf{WTQ dataset} \\
			\hline
			{\# of tables}  & {273,816} & 2,108\\
			\hline	
			{Avg \# of columns}  & {4.55} & 6.38\\
			{Max \# of columns}  & {52} & 25\\
			{Min \# of columns}  & {1} &  3 \\
			\hline
			{Avg \# of rows} & {9.15} & 28.50 \\
			{Max \# of rows} & {1,517} & 754\\
			{Min \# of rows} & {2} & 5\\
			\hline
			{\# of questions}  & {21,113} & 22,033\\
			{Avg \# of questions}  & {4.61} & 11.25\\					
			\hline
		\end{tabular}
		\caption{Statistics of WebQueryTable (WQT) dataset and WikiTableQuestions  (WTQ) dataset.}
		\label{table:statistic}
	\end{small}
\end{table}

We sampled 200 examples to analyze the distribution of the query types in our dataset.
We observe that 69.5\% queries are asking about ``a list of XXX'', such as ``\textit{list of countries and capitals}'' and ``\textit{major cities in netherlands}", and about 24.5\% queries are asking about an attribute of an object, such as
``\textit{density of liquid water temperature}''.
We randomly separate the dataset as training, validation, test with a 70:10:20 split.

We also conduct a synthetic experiment for table retrieval on
\textbf{WikiTableQuestions} \cite{pasupat2015wtq}, which is a widely used dataset for table-based question answering.
It contains 2,108 HTML tables extracted from Wikipedia.
Workers from Amazon Mechanical Turk are asked to write several relevant questions for each table.
Since each query is written for a specific table, we believe that each pair of query-table can also be used as an instance for table retrieval.
The difference between WikiTableQuestions and WebQueryTable is that the questions in WikiTableQuestions mainly focus on the local regions, such as cells or columns, of a table while the queries in WebQueryTable mainly focus on the global content of a table.
The number of table index in WikiTableQuestions is 2,108, which is smaller than the number of table index in WebQueryTable.
We randomly split the 22,033 question-table pairs into training (70\%),  development (10\%) and test (20\%).

In the candidate table retrieval phase, we encode a table as bag-of-words to guarantee the efficiency of the approach. 
Specifically, on WebQueryTable dataset we represent a table with caption and headers. On WikiTableQuestions dataset we represent a table with caption, headers and cells.
The recalls of the candidate table retrieval step on WikiTableQuestions and WebQueryTable datasets are 56.91\% and 69.57\%, respectively.
The performance of table ranking is evaluated with \textit{Mean Average Precision (MAP)} and \textit{Precision@1 (P@1)} \cite{manning2008ir}.
When evaluating the performance on table ranking, 
we filter out the following special cases that 
only one candidate table is returned or 
the correct answer is not contained in the retrieved tables in the first step.
Hyper parameters are tuned on the validation set.

\subsection{Results on WebQueryTable}
Table \ref{results:webquerytable} shows the performance of different approaches on the WebQueryTable dataset.

\begin{table}[h]
	\centering
	\begin{tabular}{l|c|c}
		\hline
		Setting & MAP & P@1\\
		\hline
		BM25 & 58.23 & 47.12\\
		Feature & 61.02 & 47.79\\
		NeuralNet & 61.94 & 49.02 \\
		Feature + NeuralNet & 67.18 & 54.15\\
		\hline
	\end{tabular}
	\caption{Results on the WebQueryTable dataset.}
	\label{results:webquerytable}
\end{table}

We compare between different features for table ranking.
An intuitive baseline is to represent a table as bag-of-words, represent a query with bag-of-words, and calculate their similarity with cosine similarity.
Therefore, we use the BM25 score which is calculated in the candidate table retrieval step. 
This baseline is abbreviated as \textbf{BM25}. 
We also report the results of using designed features (\textbf{Feature}) described in Section \ref{section:features} and neural networks (\textbf{NeuralNet}) described in Section~\ref{section:nn}.
Results from Table \ref{results:webquerytable} show that the neural networks perform comparably with the designed features, and obtain better performance than the BM25 baseline. 
This results reflect the necessary of taking into account the table structure for table retrieval.
Furthermore, we can find that combining designed features and neural networks could achieve further improvement, which indicates the complementation between them.  

We further investigate the effects of headers, cells and caption for table retrieval on WebQueryTable.
We first use each aspect separately and then increasingly combine different aspects.  
Results are given in Table \ref{results:webquerytable-detail}.
We can find that in general the performance of an aspect in designed features is consistent with its performance in neural networks. 
Caption is the most effective aspect on WebQueryTable. 
This is reasonable as we find that majority of the queries are asking about a list of objects, such as \textit{``polish rivers"}, \textit{``world top 5 mountains"} and \textit{``list of american cruise lines"}. 
These intentions are more likely to be matched in the caption of a table.
Combining more aspects could get better results. Using cells, headers and caption simultaneously gets the best results. 

\begin{table}[h]
	\centering
	\begin{tabular}{l|c c|c c}
		\hline
		\multirow{2}{2cm}{Setting} & \multicolumn{2}{|c|}{Feature} & \multicolumn{2}{|c}{NeuralNet} \\
		\cline{2-5}
		& MAP & P@1 & MAP & P@1 \\
		\hline			
		Header (H) & 22.39 & 9.76 & 26.03 & 13.35 \\
		Cell (Cel) & 28.85 & 14.95 & 27.47 & 12.92 \\
		Caption (Cap) & 57.12 & 56.83 & 60.16 & 48.48 \\
		\hline
		H + Cel & 31.99 & 17.08 & 30.73 & 16.25 \\
		H + Cel + Cap & 61.02 & 47.79 & 61.94 & 49.02 \\
		\hline
	\end{tabular}
	\caption{Performance on WebQueryTable dataset with different aspects.}
	\label{results:webquerytable-detail}
\end{table}

\begin{figure}[h]
	\centering
	\includegraphics[width=0.5\textwidth]
	{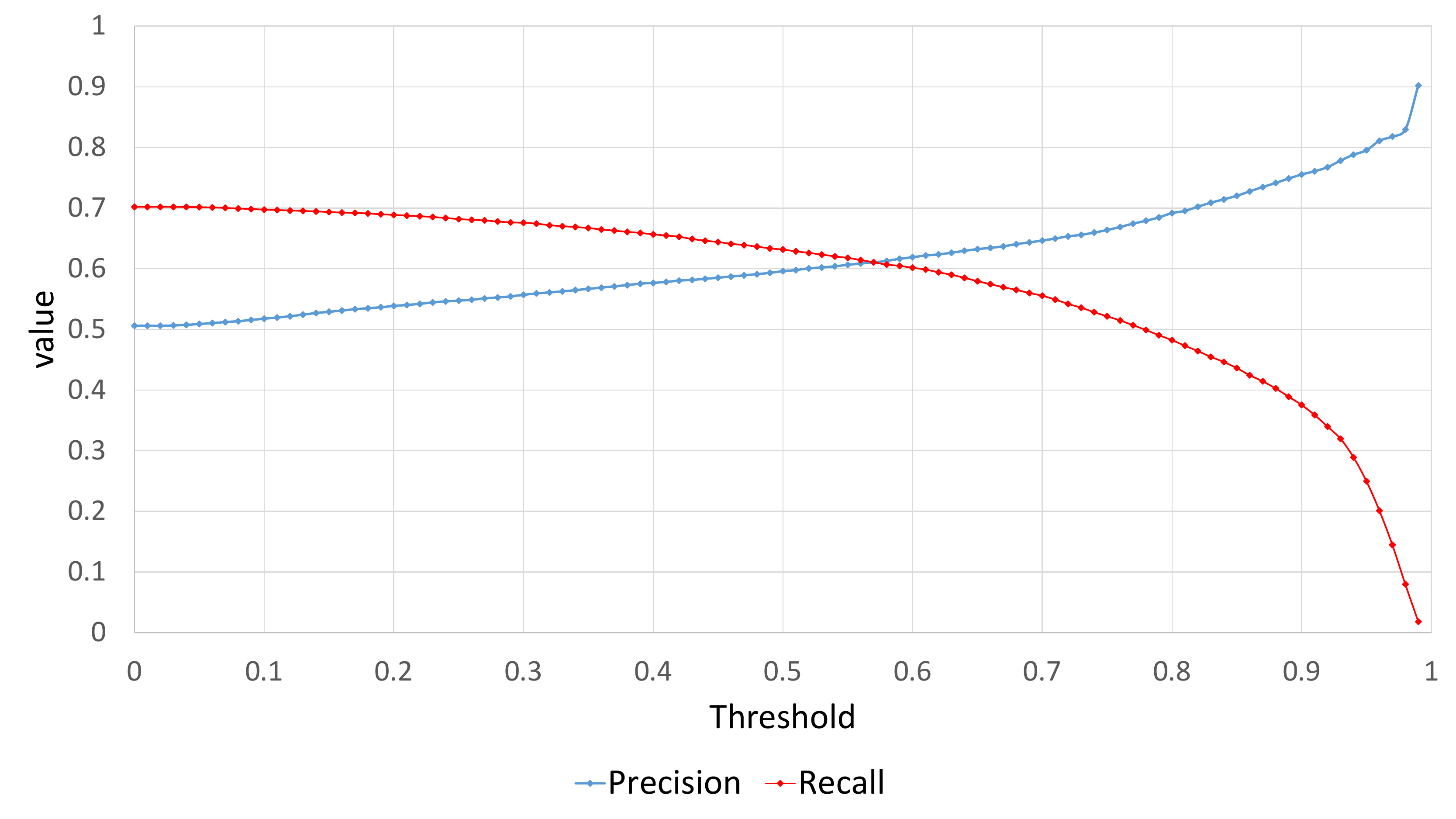}
	\caption{PR Curve on WebQueryTable.}
	\label{figure:pr}
\end{figure}

Moreover, we investigate whether using a higher threshold could obtain a better precision. 
Therefore, we increasingly use a set of thresholds, and calculate the corresponding precision and recall in different conditions.
An instance is considered to be correct if the top ranked table is correct and its ranking score is greater than the threshold.
Results of our NeuralNet approach on WebQueryTable are given in \ref{figure:pr}. We can see that using larger threshold results in lower recall and higher precision. The results are consistent with our intuition.

\begin{figure*}[t]
	\centering
	\includegraphics[width=1\textwidth]
	{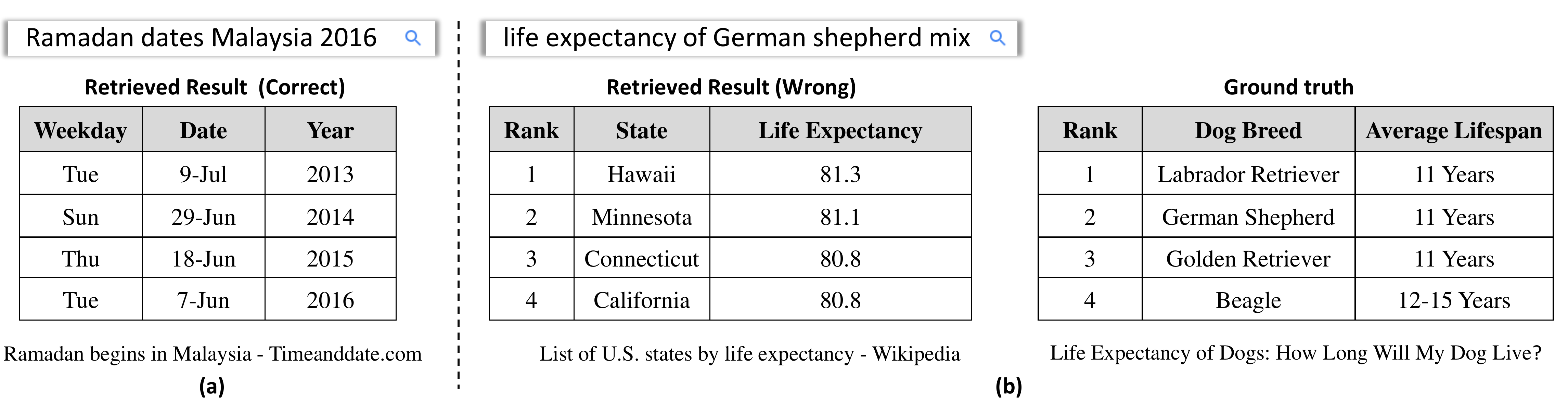}
	\caption{Results generated by NeuralNet on WebQueryTable.}
	\label{figure:case-study}
\end{figure*}

\begin{figure}[h]
	\centering
	\includegraphics[width=0.48\textwidth]
	{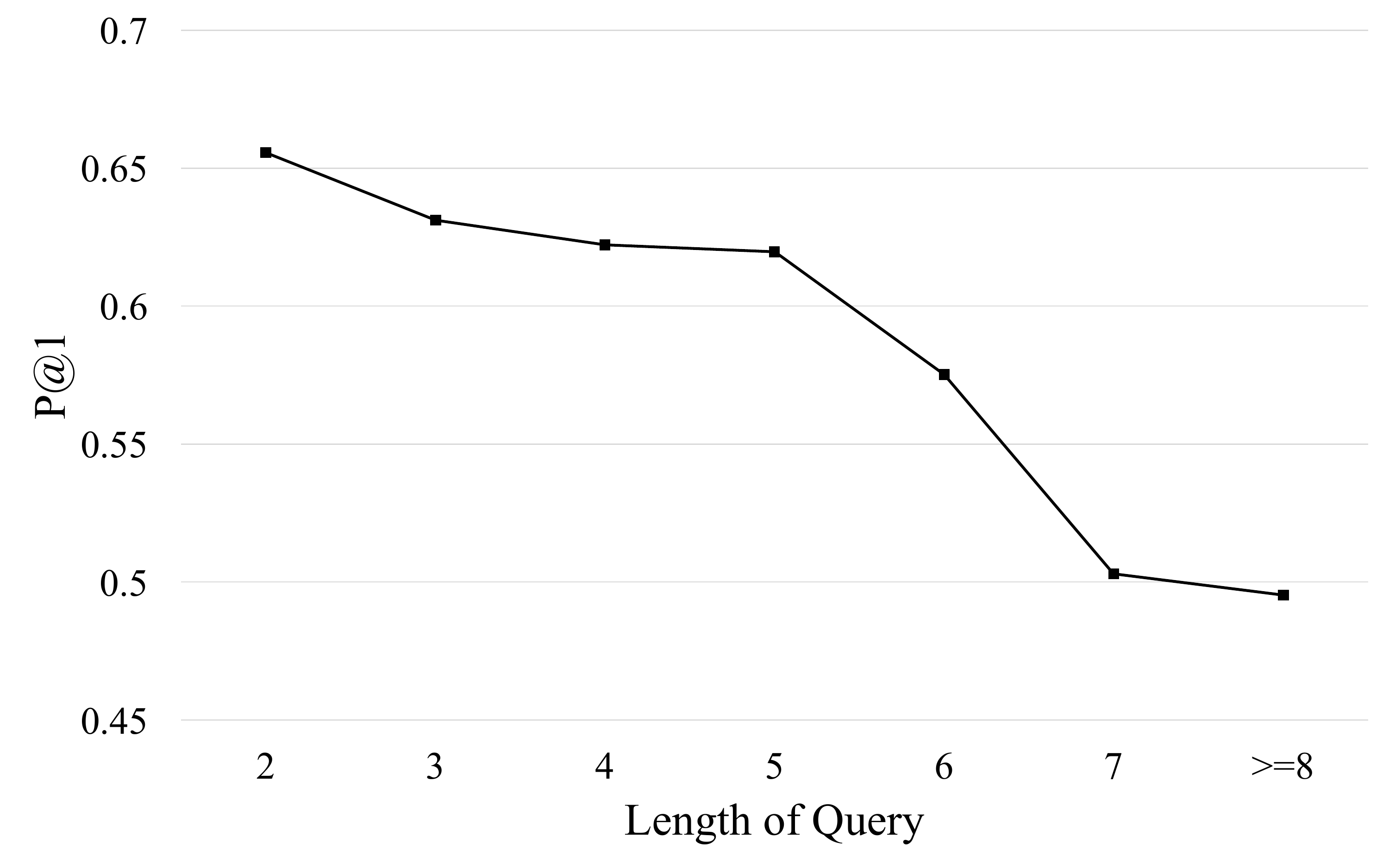}
	\caption{P@1 with different query length on WebQueryTable dataset.}
	\label{figure:length}
\end{figure}

We conduct case study on our NeuralNet approach and find that the performance is sensitive to the length of queries.
Therefore, we split the test set to several groups according to the length of queries. 
Results are given in Figure \ref{figure:length}.
We can find that the performance of the approach decreases with the increase of query length.
When the query length changes from 6 to 7, the performance of P@1 decreases rapidly from 58.12\% to 50.23\%.
Through doing case study, we find that long queries contain more word dependencies.
Therefore, having a good understanding about the intention of a query requires deep query understanding. 
Leveraging external knowledge to connect query and table is a potential solution to deal with long queries.

We illustrate two examples generated by our NeuralNet approach in Figure \ref{figure:case-study}.
The example in Figure \ref{figure:case-study}(a) is a satisfied case that the top ranked result is the correct answer.
We can find that the model uses evidences from different aspects to match between a query and a table. 
In this example, the supporting evidences come from caption (``\textit{ramadan}" and ``\textit{malaysia}"), headers (``\textit{dates}") and cells (``\textit{2016}").
The example in Figure~\ref{figure:case-study}(b) is a dissatisfied case.
We can find that the top ranked result contains ``\textit{life expectancy}" in both caption and header, however, it is talking about the people in U.S. rather than ``\textit{german shepherd}". 
Despite the correct table contains a cell whose content is ``\textit{german shepherd}", it still does not obtain a higher rank than the left table. The reason might be that the weight for header is larger than the weight for cells. 

\subsection{Results on WikiTableQuestions}

Table \ref{results:wikitablequestion} shows the results of table ranking on the WikiTableQuestions dataset.

\begin{table}[h]
	\centering
	\begin{tabular}{l|c|c}
		\hline
		Setting & MAP & P@1\\
		\hline
		BM25 & 51.02 & 41.02\\
		CDSSM-Header & 45.92 & 31.58\\
		Feature & 67.70 & 56.25\\
		NeuralNet & 67.44 & 54.95 \\
		Feature + NeuralNet & 72.49 & 61.50\\
		\hline
	\end{tabular}
	\caption{Results on the WikiTableQuestions dataset with different features.}
	\label{results:wikitablequestion}
\end{table}
We implement two baselines. 
The first baseline is BM25, which is the same baseline we have used for comparison on the WebQueryTable dataset.
The second baseline is header grounding, which is partly inspired by \newcite{VLDB2011GG} who show the effectiveness of the semantic relationship between query and table header.
We implement a CDSSM \cite{shen2014CDSSM} approach to match between a table header and a query.
We train the model by minimizing the cross-entropy error, where the ground truth is the header of the answer.
Results are given in Table \ref{results:wikitablequestion}. We can find that designed features perform comparably with neural networks, and both of them perform better than BM25 and column grounding baselines.
Combining designed features and neural networks obtains further improvement.

We also study the effects of different aspects on the WikiTableQuestions dataset.
Results are given in Table \ref{results:wikitablequestion-detail}. 
\begin{table}[h]
	\centering
	\begin{tabular}{l|cc|cc}
		\hline
		\multirow{2}{2cm}{Setting} & \multicolumn{2}{|c|}{Feature} & \multicolumn{2}{|c}{NeuralNet} \\
		\cline{2-5}
		& MAP & P@1 & MAP & P@1 \\
		\hline			
		Header (H) & 46.36 & 32.52 & 52.93 & 36.47 \\
		Cell (Cel) & 44.33 & 30.97 & 43.49 & 26.41 \\
		Caption (Cap) & 33.36 & 24.79 & 46.83 & 31.54 \\			
		\hline			
		H + Cel & 60.03 & 47.42 & 60.55 & 45.71 \\
		H + Cel + Cap & 67.70 & 56.25 & 67.44 & 54.95 \\
		\hline
	\end{tabular}
	\caption{Results on the WikiTableQuestions dataset with different aspects.}
	\label{results:wikitablequestion-detail}
\end{table}
We can find that the effects of different aspect in designed features and neural networks are consistent.
Using more aspects could achieve better performance.
Using all aspects obtains the best performance.
We also find that the most effective aspect for WikiTableQuestions is header. This is different from the phenomenon in WebQueryTable that the most effective aspect is caption.
We believe that this is because the questions in WikiTableQuestions typically include content constrains from cells or headers. 
Two randomly sampled questions are \textit{``which country won the 1994 europeans men's handball championship's preliminary round?"} and \textit{``what party had 7,115 inactive voters as of october 25, 2005?"}.
On the contrary, queries from WebTableQuery usually do not use information from specific headers or cells.
Examples include \textit{``polish rivers"}, \textit{``world top 5 mountains"} and \textit{``list of american cruise lines"}.
From Table \ref{table:statistic}, we can also find that the question in WikiTableQuestions are longer than the queries in WebQueryTable. 
In addition, we observe that not all the questions from WikiTableQuestions are suitable for table retrieval. 
An example is \textit{``what was the first player to be drafted in this table?"}. 

\section{Related Work}
Our work connects to the fields of database and natural language processing.

There exists several works in database community that aims at finding related tables from keyword queries.
A representative work is given by 
\newcite{VLDB2008GG}, 
which considers table search as a special case of document search task and represent a table with its surrounding text and page title.
\newcite{VLDB2010india} use YAGO ontology to annotate tables with column and relationship labels.
\newcite{VLDB2011GG} go one step further and use labels and relationships extracted from the web.
\newcite{VLDB2012IBM} focus on the queries that describe table columns, and retrieve tables based on column mapping.
%
%
%
There also exists table-related studies such as searching related tables from a table \cite{SIGMOD2012GG},  assembling a table from list in web page \cite{VLDB2009india} and extracting tables using tabular structure from web page \cite{WWW2007web2table}.
Our work differs from this line of research in that we focus on exploring the content of table to find relevant tables from web queries. 

Our work relates to a line of research works that learn continuous representation of structured knowledge with neural network for natural language processing tasks.
For example, \newcite{neelakantan2015neural,pengcheng2015} develop neural operator on the basis of table representation and
apply the model to question answering.
\newcite{yin2015NGQA} introduce a KB-enhanced sequence-to-sequence approach that generates natural language answers to simple factoid questions based on facts from KB.
\newcite{mei-bansal-walter:2016:N16-1} develop a LSTM based recurrent neural network to generate natural language weather forecast and  sportscasting commentary from database records.
\newcite{serban-EtAl:2016:P16-1} introduce a recurrent neural network approach, which takes fact representation as input and generates factoid question from a fact from Freebase.
\newcite{table2textEMNLP2016} presented an neural language model that generates biographical sentences from Wikipedia infobox.

Our neural network approach relates to the recent advances of attention mechanism and reasoning over external memory in artificial intelligence \cite{Bahdanau2015,Sukhbaatar2015e2emn,graves2016hybrid}. 
Researchers typically represent a memory as a continuous vector or matrix, and develop neural network based controller, reader and writer to reason over the memory.
The memory could be addressed by a ``soft'' attention mechanism trainable by standard back-propagation methods or a ``hard'' attention mechanism trainable by REINFORCE \cite{williams1992simple}.
In this work, we use the soft attention mechanism, which could be easily optimized and has been successfully applied in nlp tasks \cite{Bahdanau2015,Sukhbaatar2015e2emn}.


\section{Conclusion}
In this paper, we give an empirical study of content-based table retrieval for web queries.
We implement a feature-based approach and a neural network based approach, and release a new dataset consisting of web queries and web tables.
We conduct comprehensive experiments on two datasets.
Results not only verify the effectiveness of our approach, but also present future challenges for content-based table retrieval.


\bibliography{acl2017}

\begin{thebibliography}{}
\expandafter\ifx\csname natexlab\endcsname\relax\def\natexlab#1{#1}\fi

\bibitem[{Bahdanau et~al.(2015)Bahdanau, Cho, and Bengio}]{Bahdanau2015}
Dzmitry Bahdanau, Kyunghyun Cho, and Yoshua Bengio. 2015.
\newblock Neural machine translation by jointly learning to align and
  translate.
\newblock {\em International Conference on Learning Representations (ICLR)\/} .

\bibitem[{Balakrishnan et~al.(2015)Balakrishnan, Halevy, Harb, Lee, Madhavan,
  Rostamizadeh, Shen, Wilder, Wu, and Yu}]{CIDR2015GG}
Sreeram Balakrishnan, Alon~Y Halevy, Boulos Harb, Hongrae Lee, Jayant Madhavan,
  Afshin Rostamizadeh, Warren Shen, Kenneth Wilder, Fei Wu, and Cong Yu. 2015.
\newblock Applying webtables in practice.
\newblock {\em Proceedings of Conference on Innovative Data Systems Research
  (CIDR)\/} .

\bibitem[{Burges(2010)}]{mart2010l2r}
Christopher~JC Burges. 2010.
\newblock From ranknet to lambdarank to lambdamart: An overview.
\newblock {\em Microsoft Research Technical Report MSR-TR-2010-82\/}
  11(23-581):81.

\bibitem[{Cafarella et~al.(2008)Cafarella, Halevy, Wang, Wu, and
  Zhang}]{VLDB2008GG}
Michael~J Cafarella, Alon Halevy, Daisy~Zhe Wang, Eugene Wu, and Yang Zhang.
  2008.
\newblock Webtables: exploring the power of tables on the web.
\newblock {\em Proceedings of the VLDB Endowment\/} 1(1):538--549.

\bibitem[{Cho et~al.(2014)Cho, van Merrienboer, Gulcehre, Bahdanau, Bougares,
  Schwenk, and Bengio}]{cho-EtAl:2014:EMNLP2014}
Kyunghyun Cho, Bart van Merrienboer, Caglar Gulcehre, Dzmitry Bahdanau, Fethi
  Bougares, Holger Schwenk, and Yoshua Bengio. 2014.
\newblock \href{http://www.aclweb.org/anthology/D14-1179}{Learning phrase
  representations using rnn encoder--decoder for statistical machine
  translation}.
\newblock In {\em Proceedings of the 2014 Conference on Empirical Methods in
  Natural Language Processing (EMNLP)\/}. Association for Computational
  Linguistics, Doha, Qatar, pages 1724--1734.
\newblock
  \href{http://www.aclweb.org/anthology/D14-1179}{http://www.aclweb.org/anthology/D14-1179}.

\bibitem[{Chung et~al.(2014)Chung, Gulcehre, Cho, and
  Bengio}]{chung2014empirical}
Junyoung Chung, Caglar Gulcehre, KyungHyun Cho, and Yoshua Bengio. 2014.
\newblock Empirical evaluation of gated recurrent neural networks on sequence
  modeling.
\newblock {\em arXiv preprint arXiv:1412.3555\/} .

\bibitem[{Das~Sarma et~al.(2012)Das~Sarma, Fang, Gupta, Halevy, Lee, Wu, Xin,
  and Yu}]{SIGMOD2012GG}
Anish Das~Sarma, Lujun Fang, Nitin Gupta, Alon Halevy, Hongrae Lee, Fei Wu,
  Reynold Xin, and Cong Yu. 2012.
\newblock Finding related tables.
\newblock In {\em Proceedings of the 2012 ACM SIGMOD International Conference
  on Management of Data\/}. ACM, pages 817--828.

\bibitem[{Fader et~al.(2013)Fader, Zettlemoyer, and
  Etzioni}]{Fader13WikiAnswer}
Anthony Fader, Luke Zettlemoyer, and Oren Etzioni. 2013.
\newblock Paraphrase-driven learning for open question answering.
\newblock In {\em Proceedings of Annual Meeting of the Association for
  Computational Linguistics (ACL)\/}.

\bibitem[{Gatterbauer et~al.(2007)Gatterbauer, Bohunsky, Herzog, Kr{\"u}pl, and
  Pollak}]{WWW2007web2table}
Wolfgang Gatterbauer, Paul Bohunsky, Marcus Herzog, Bernhard Kr{\"u}pl, and
  Bernhard Pollak. 2007.
\newblock Towards domain-independent information extraction from web tables.
\newblock In {\em Proceedings of the 16th international conference on World
  Wide Web (WWW)\/}. ACM, pages 71--80.

\bibitem[{Graves et~al.(2016)Graves, Wayne, Reynolds, Harley, Danihelka,
  Grabska-Barwi{\'n}ska, Colmenarejo, Grefenstette, Ramalho, Agapiou
  et~al.}]{graves2016hybrid}
Alex Graves, Greg Wayne, Malcolm Reynolds, Tim Harley, Ivo Danihelka, Agnieszka
  Grabska-Barwi{\'n}ska, Sergio~G{\'o}mez Colmenarejo, Edward Grefenstette,
  Tiago Ramalho, John Agapiou, et~al. 2016.
\newblock Hybrid computing using a neural network with dynamic external memory.
\newblock {\em Nature\/} 538(7626):471--476.

\bibitem[{Gupta and Sarawagi(2009)}]{VLDB2009india}
Rahul Gupta and Sunita Sarawagi. 2009.
\newblock Answering table augmentation queries from unstructured lists on the
  web.
\newblock {\em Proceedings of the VLDB Endowment\/} 2(1):289--300.

\bibitem[{Koehn et~al.(2003)Koehn, Och, and Marcu}]{koehn2003PP}
Philipp Koehn, Franz~Josef Och, and Daniel Marcu. 2003.
\newblock Statistical phrase-based translation.
\newblock {\em Proceedings of Annual Conference of the North American Chapter
  of the Association for Computational Linguistics: Human Language Technologies
  (NAACL-HLT)\/} 1:48--54.

\bibitem[{Lebret et~al.(2016)Lebret, Grangier, and Auli}]{table2textEMNLP2016}
R\'{e}mi Lebret, David Grangier, and Michael Auli. 2016.
\newblock Neural text generation from structured data with application to the
  biography domain.
\newblock In {\em Proceedings of the 2016 Conference on Empirical Methods in
  Natural Language (EMNLP)\/}.

\bibitem[{Limaye et~al.(2010)Limaye, Sarawagi, and Chakrabarti}]{VLDB2010india}
Girija Limaye, Sunita Sarawagi, and Soumen Chakrabarti. 2010.
\newblock Annotating and searching web tables using entities, types and
  relationships.
\newblock {\em Proceedings of the VLDB Endowment\/} 3(1-2):1338--1347.

\bibitem[{Manning et~al.(2008)Manning, Raghavan, Sch{\"u}tze
  et~al.}]{manning2008ir}
Christopher~D Manning, Prabhakar Raghavan, Hinrich Sch{\"u}tze, et~al. 2008.
\newblock {\em Introduction to information retrieval\/}, volume~1.
\newblock Cambridge university press Cambridge.

\bibitem[{Mei et~al.(2016)Mei, Bansal, and
  Walter}]{mei-bansal-walter:2016:N16-1}
Hongyuan Mei, Mohit Bansal, and Matthew~R. Walter. 2016.
\newblock \href{http://www.aclweb.org/anthology/N16-1086}{What to talk about
  and how? selective generation using lstms with coarse-to-fine alignment}.
\newblock In {\em Proceedings of the 2016 Conference of the North American
  Chapter of the Association for Computational Linguistics: Human Language
  Technologies\/}. Association for Computational Linguistics, San Diego,
  California, pages 720--730.
\newblock
  \href{http://www.aclweb.org/anthology/N16-1086}{http://www.aclweb.org/anthology/N16-1086}.

\bibitem[{Mikolov et~al.(2013)Mikolov, Sutskever, Chen, Corrado, and
  Dean}]{mikolov2013w2v}
Tomas Mikolov, Ilya Sutskever, Kai Chen, Greg~S Corrado, and Jeff Dean. 2013.
\newblock Distributed representations of words and phrases and their
  compositionality.
\newblock In {\em Advances in neural information processing systems (NIPS)\/}.
  pages 3111--3119.

\bibitem[{Neelakantan et~al.(2015)Neelakantan, Le, and
  Sutskever}]{neelakantan2015neural}
Arvind Neelakantan, Quoc~V Le, and Ilya Sutskever. 2015.
\newblock Neural programmer: Inducing latent programs with gradient descent.
\newblock {\em arXiv preprint arXiv:1511.04834\/} .

\bibitem[{Pasupat and Liang(2015)}]{pasupat2015wtq}
Panupong Pasupat and Percy Liang. 2015.
\newblock Compositional semantic parsing on semi-structured tables.
\newblock {\em Proceedings of Annual Meeting of the Association for
  Computational Linguistics (ACL)\/} .

\bibitem[{Pimplikar and Sarawagi(2012)}]{VLDB2012IBM}
Rakesh Pimplikar and Sunita Sarawagi. 2012.
\newblock Answering table queries on the web using column keywords.
\newblock {\em Proceedings of the VLDB Endowment\/} 5(10):908--919.

\bibitem[{Robertson et~al.(1995)Robertson, Walker, Jones, Hancock-Beaulieu,
  Gatford et~al.}]{bm25}
Stephen~E Robertson, Steve Walker, Susan Jones, Micheline~M Hancock-Beaulieu,
  Mike Gatford, et~al. 1995.
\newblock Okapi at trec-3.
\newblock {\em NIST SPECIAL PUBLICATION SP\/} 109:109.

\bibitem[{Serban et~al.(2016)Serban, Garc\'{i}a-Dur\'{a}n, Gulcehre, Ahn,
  Chandar, Courville, and Bengio}]{serban-EtAl:2016:P16-1}
Iulian~Vlad Serban, Alberto Garc\'{i}a-Dur\'{a}n, Caglar Gulcehre, Sungjin Ahn,
  Sarath Chandar, Aaron Courville, and Yoshua Bengio. 2016.
\newblock \href{http://www.aclweb.org/anthology/P16-1056}{Generating factoid
  questions with recurrent neural networks: The 30m factoid question-answer
  corpus}.
\newblock In {\em Proceedings of the 54th Annual Meeting of the Association for
  Computational Linguistics (Volume 1: Long Papers)\/}. Association for
  Computational Linguistics, Berlin, Germany, pages 588--598.
\newblock
  \href{http://www.aclweb.org/anthology/P16-1056}{http://www.aclweb.org/anthology/P16-1056}.

\bibitem[{Shen et~al.(2014)Shen, He, Gao, Deng, and Mesnil}]{shen2014CDSSM}
Yelong Shen, Xiaodong He, Jianfeng Gao, Li~Deng, and Gr{\'e}goire Mesnil. 2014.
\newblock A latent semantic model with convolutional-pooling structure for
  information retrieval.
\newblock In {\em Proceedings of the Conference on Information and Knowledge
  Management (CIKM)\/}. pages 101--110.

\bibitem[{Sukhbaatar et~al.(2015)Sukhbaatar, Szlam, Weston, and
  Fergus}]{Sukhbaatar2015e2emn}
Sainbayar Sukhbaatar, Arthur Szlam, Jason Weston, and Rob Fergus. 2015.
\newblock End-to-end memory networks.
\newblock In {\em Advances in Neural Information Processing Systems (NIPS)\/}.
  pages 2431--2439.

\bibitem[{Venetis et~al.(2011)Venetis, Halevy, Madhavan, Pa{\c{s}}ca, Shen, Wu,
  Miao, and Wu}]{VLDB2011GG}
Petros Venetis, Alon Halevy, Jayant Madhavan, Marius Pa{\c{s}}ca, Warren Shen,
  Fei Wu, Gengxin Miao, and Chung Wu. 2011.
\newblock Recovering semantics of tables on the web.
\newblock {\em Proceedings of the VLDB Endowment\/} 4(9):528--538.

\bibitem[{Vinyals et~al.(2015)Vinyals, Bengio, and Kudlur}]{vinyals2015order}
Oriol Vinyals, Samy Bengio, and Manjunath Kudlur. 2015.
\newblock Order matters: Sequence to sequence for sets.
\newblock {\em arXiv preprint arXiv:1511.06391\/} .

\bibitem[{Williams(1992)}]{williams1992simple}
Ronald~J Williams. 1992.
\newblock Simple statistical gradient-following algorithms for connectionist
  reinforcement learning.
\newblock {\em Machine learning\/} 8(3-4):229--256.

\bibitem[{Yin et~al.(2015{\natexlab{a}})Yin, Jiang, Lu, Shang, Li, and
  Li}]{yin2015NGQA}
Jun Yin, Xin Jiang, Zhengdong Lu, Lifeng Shang, Hang Li, and Xiaoming Li.
  2015{\natexlab{a}}.
\newblock Neural generative question answering.
\newblock {\em arXiv preprint arXiv:1512.01337\/} .

\bibitem[{Yin et~al.(2015{\natexlab{b}})Yin, Lu, Li, and Kao}]{pengcheng2015}
Pengcheng Yin, Zhengdong Lu, Hang Li, and Ben Kao. 2015{\natexlab{b}}.
\newblock Neural enquirer: Learning to query tables with natural language.
\newblock {\em arXiv preprint arXiv:1512.00965\/} .

\end{thebibliography}
\bibliographystyle{emnlp_natbib}

\end{document}